\documentclass[letterpaper, 10 pt, conference]{ieeeconf}  

\IEEEoverridecommandlockouts                              

\usepackage{amsmath,amssymb,amsfonts}
\usepackage{algpseudocode,algorithm}
\usepackage{subcaption}
\usepackage{graphicx}
\usepackage{cite}
\usepackage{url}

\usepackage[T1]{fontenc}

\title{\LARGE \bf
Intentionally-underestimated Value Function at Terminal State for Temporal-difference Learning with Mis-designed Reward
}

\author{Taisuke Kobayashi$^{1}$
\thanks{$^{1}$Taisuke Kobayashi is with National Institute of Informatics (NII) and with The Graduate University for Advanced Studies (SOKENDAI), 2-1-2 Hitotsubashi, Chiyoda-ku, Tokyo, 101-8430, Japan.
        {\tt\small kobayashi@nii.ac.jp}}%
}

\begin{document}

\maketitle
\thispagestyle{empty}
\pagestyle{empty}

\begin{abstract}

Robot control using reinforcement learning has become popular, but its learning process generally terminates halfway through an episode for safety and time-saving reasons.
This study addresses the problem of the most popular exception handling that temporal-difference (TD) learning performs at such termination.
That is, by forcibly assuming zero value after termination, unintentionally implicit underestimation or overestimation occurs, depending on the reward design in the normal states.
When the episode is terminated due to task failure, the failure may be highly valued with the unintentional overestimation, and the wrong policy may be acquired.
Although this problem can be avoided by paying attention to the reward design, it is essential in practical use of TD learning to review the exception handling at termination.
This paper therefore proposes a method to intentionally underestimate the value after termination to avoid learning failures due to the unintentional overestimation.
In addition, the degree of underestimation is adjusted according to the degree of stationarity at termination, thereby preventing excessive exploration due to the intentional underestimation.
Simulations and real robot experiments showed that the proposed method can stably obtain the optimal policies for various tasks and reward designs.

\end{abstract}

%
%
\section{Introduction}

Recently, reinforcement learning (RL)~\cite{sutton2018reinforcement} has attracted much attention as a control technique for complex systems, such as robots.
Various RL algorithms have been proposed, and they can be mainly divided into Monte Carlo methods (e.g.~\cite{miyashita2018mirror}) and methods that utilize temporal difference (TD) errors, so-called TD learning (e.g.~\cite{mnih2015human,schulman2017proximal,haarnoja2018soft,fujimoto2018addressing,tsurumine2019deep,vieillard2020munchausen}).
Theoretically, these two types of methods have advantages and disadvantages due to the bias-variance trade-off problem, but TD learning has become more popular in recent years, perhaps due to its ease of use in conjunction with experience replay~\cite{lin1992self}.
In fact, the major RL algorithms (e.g. DQN~\cite{mnih2015human}, SAC~\cite{haarnoja2018soft}, etc.) seem to be based on TD learning and have been applied to various tasks.

When acquiring a robot controller through TD learning, the robot faces various experiences, which are utilized for learning, by repeating the target task many times through trial-and-error.
In this case, the episode, in which the task is attempted, is terminated at an appropriate timing from the viewpoint of safety, time constraint, and so on, then retrying the task from the beginning as a new episode.
If the length of such a finite episode is always constant, it is sufficient to consider the sum of rewards up to a specified time in the future, so this setting was often employed until a while ago~\cite{kober2013reinforcement}.
In practice, however, there are various reasons for terminating an episode (e.g. to guarantee robot safety, to try the task with other conditions), so in recent years, TD learning considers the future infinitely while discounting it~\cite{ibarz2021train}.
Note that ``model-based'' RL still employs the former setting frequently in combination with finite-length future prediction~\cite{polydoros2017survey}.

This study focuses on exception handling at the end of each episode in TD learning.
This is activated when the tackled task is on the success or failure (or its time limit is over).
In that time, although TD learning requires the value of the next time step to learn the value (and policy) function of the current time step, the value after termination cannot be correctly derived.
Therefore, the value after termination must be assumed.
In the recent libraries~\cite{liang2018rllib,raffin2021stable,huang2022cleanrl}, it is generally given as zero so that it is uninformative.

\begin{figure}[tb]
    \centering
    \includegraphics[keepaspectratio=true,width=0.96\linewidth]{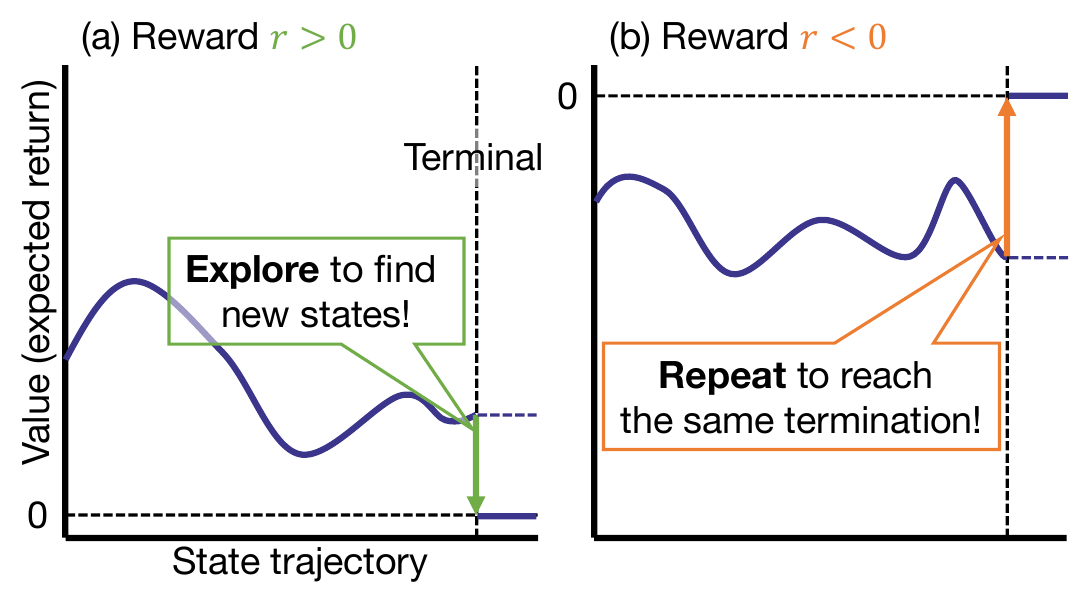}
    \caption{Implicit underestimation or overestimation}
    \label{fig:img_under_over}
\end{figure}

However, such an exception handling is uninformative only if the reward function is designed to be zero on average.
This assumption is not realistic, and in most cases, the reward function is biased either positively or negatively.
In that time, the zero value after termination will be underestimated or overestimated compared to the true (but implicit) value after termination (see Fig.~\ref{fig:img_under_over}).
The underestimation would be allowable with little risk of learning failure, since it encourages exploration to find new endpoints rather than accepting the one visited.
On the other hand, the overestimation may lead to a collapse of learning, since it prefers the endpoint visited, which may mean failure of the task, rather than new ones.
As a result, even with the state-of-the-art RL algorithm, RL will never learn the task unless users properly design the corresponding reward function.
However, it would be a terrible idea to ask all users who are not familiar with the details of RL to design the reward function properly while considering the above caveats in mind.

In this study, instead of the currently most popular exception handling with the zero value after termination, a novel exception handling, which intentionally always underestimates the value after termination, is proposed.
Specifically, the state after termination is regarded to be an absorbing state~\cite{feller1991introduction}, which always returns a certain amount of reward without state transition.
The state just before termination is assumed to converge to this absorbing state ``gradually''.
With this assumption, the term that represents the difference from the normal TD learning is revealed.
By heuristically shaping it, the proposed method always underestimates the value after termination, facilitating exploration implicitly.
In addition, the degree of underestimation is automatically reduced depending on the stationarity at termination in order to mitigate the excessive exploration.

In numerical simulations~\cite{todorov2012mujoco,coumans2016pybullet,tunyasuvunakool2020dm_control}, the learning performance of the conventional method (where the value after termination is assumed to be zero) significantly degrades depending on the reward design.
On the other hand, when the proposed method is incorporated into multiple RL algorithms based on TD learning~\cite{kobayashi2023reward,kobayashi2023soft}, it solves various tasks, independent of the reward design and the used algorithms.
It is also experimentally verified that the proposed method can acquire the tasks with the standard reward design more stably than the conventional methods.
Finally, the proposed method is empilified through real robot experiments for synchronized swinging motion with variable stiffness actuators (VSAs)~\cite{catalano2011vsa}.

The contributions of this paper are threefold:
\begin{enumerate}
    \item It has been first pointed out that the conventional exception handling at termination collapses learning depending on the reward design.
    \item The new exception handling at termination has been designed to intentionally promotes exploration without trapping into local optima.
    \item The proposed method has been empirically verified using multiple RL algorithms and tasks.
\end{enumerate}
Although the design of the proposed method may leave room for improvement, as described later, the issue pointed out in this paper and the guideline for its solution are expected to promote the spread of TD learning to novices.

\section{Preliminaries}

\subsection{Temporal-difference learning}

RL~\cite{sutton2018reinforcement} is an optimal control problem to find the optimal policy $\pi^\ast$ under Markov decision process represented by the tuple $(\mathcal{S}, \mathcal{A}, \mathcal{R}, p_e, p_0)$.
Specifically, a state $s \in \mathcal{S}$ is sampled from an unknown environment with initial state probability $p_0$ or state transition probability $p_e$, and an agent stochastically determines its action $a \in \mathcal{A}$ according to $s$ from a trainable policy $\pi(a \mid s)$.
The agent interacts with the environment using $a$ and stochastically transitions to the next state $s^\prime \in \mathcal{S}$ according to the state transition probability $p_e(s^\prime \mid s, a)$.
In this time, a reward $r \in \mathcal{R}$ is given by the environment as the criterion of this interaction.
The agent optimizes $\pi$ to maximize the sum of rewards from the current time $t$ to the infinite time future (so-called return), $R = \sum_{k=0}^\infty \gamma^k r_{t+k}$ with with $\gamma \in [0, 1)$ the discount factor.

In such an optimization problem, RL often learns the following value functions, which are the expectation of $R$.
\begin{align}
    Q(s, a) &= \mathbb{E}_{\tau \sim p_\tau}[R \mid s, a]
    \label{eq:value_action}\\
    V(s) &= \mathbb{E}_{a \sim \pi, \tau \sim p_\tau}[R \mid s] = \mathbb{E}_{a \sim \pi}[Q(s, a)]
    \label{eq:value_state}
\end{align}
where $\tau$ denotes the state-action trajectory generated from $p_\tau$ (more specifically, the product of $p_e$ and $\pi$).
$Q$ and $V$ are often learned by TD learning, which focuses on the recursiveness of $R$.
In other words, the following value is targeted for error minimization.
\begin{align}
    y =
    \begin{cases}
        r + \gamma V(s^\prime) & s^\prime \notin \mathcal{S}_\mathrm{term}
        \\
        r & s^\prime \in \mathcal{S}_\mathrm{term}
    \end{cases}
    \label{eq:td_target}
\end{align}
Here, the branching is based on whether the next state reaches the end of the episode $\mathcal{S}_\mathrm{term} \subset \mathcal{S}$.
This is because the value after the termination cannot be learned properly, then assuming $V(s^\prime \in \mathcal{S}_\mathrm{term}) = 0$ as an uninformative manner.
Although this assumption has been widely used, $V(s^\prime)$ can be calculated numerically, so this branching is sometimes ignored.
Note that the error between $y$ and the value function is so-called TD error.

\subsection{Policy improvement}

This paper demonstrates algorithms to learn the policy $\pi$ directly~\cite{schulman2017proximal,haarnoja2018soft}, and their bases are briefly introduced.
First, the optimization problem of $\pi$ to maximize $R$ (or $V$ in practice) is defined in line with the objective of RL.
\begin{align}
    \pi^\ast = \arg\max_{\pi} V(s) \ {}^\forall s \in \mathcal{S}
\end{align}
In practice, $\pi$ is approximated by some probability distribution model (e.g. Gaussian distribution) and its model parameters (and those of the function approximator that approximates them), $\theta$, are optimized.
This optimization problem is basically solved by gradient-based methods.
\begin{align}
    \theta &\gets \theta + \eta g_\theta
\end{align}
where $\eta > 0$ denotes the step size for the gradient $g_\theta$.

There are two major ways to compute $g_\theta$, and one is the policy-gradient method~\cite{williams1992simple}.
Specifically, the gradient of the objective function w.r.t. $\theta$ is derived analytically, paying attention to the fact that $V$ depends on $\pi$.
As a result, the following $g_\theta$ is obtained to update $\theta$.
\begin{align}
    g_\theta = \mathbb{E}_{s \sim p_s, a \sim \pi}[(Q(s, a) - b(s)) \nabla_{\theta} \ln \pi(a \mid s)]
    \label{eq:policy_grad1}
\end{align}
where $p_s$ denotes the probability to generate $s$, which depends on the state transition probability $p_e$ and $\pi$.
$b$ is additionally introduced as a bias to reduce the variance of $g_\theta$, and $b = V$ is frequently employed.
In addition, if $Q$ is approximated by $Q \simeq y$, i.e. $g_\theta$ is weighted by TD error, only $V$ needs to be learned without $Q$.
The first algorithm tested in this paper~\cite{kobayashi2023reward}, based on PPO~\cite{schulman2017proximal}, uses this formulation.

Another way to obtain $g_\theta$ is to use the reparameterization trick~\cite{kingma2014auto}.
By computing the gradient of the objective function w.r.t. $a$ sampled from $\pi$, $g_\theta$ is directly obtained.
\begin{align}
    g_\theta = \mathbb{E}_{s \sim p_s, \epsilon \sim p_\epsilon}[\nabla_a Q(s, a) \nabla_\theta a(\theta; \epsilon)]
    \label{eq:policy_grad2}
\end{align}
where $p_\epsilon$ denotes the probability to generate $\epsilon$, which is independent from $\theta$ but is used to sample $a$ in combination to $\theta$.
In this case, as long as $Q$ is obtained by TD learning, $\pi$ can be optimized efficiently without using $y$ for $\pi$ unlike the first case.
The second algorithm tested in this paper~\cite{kobayashi2023soft}, based on SAC~\cite{haarnoja2018soft}, employs this approach with $V(s) = \mathbb{E}_{\pi}[Q(s, a) - \alpha \ln \pi(a \mid s)]$ ($\alpha > 0$ denotes the temperature parameter), because SAC incorporates entropy maximization in its objective function.

\section{Proposal}

\subsection{Target issue}

In TD learning, where the majority of RL algorithms are classified, the value function $Q$ and/or $V$ is learned using $y$ defined in eq.~\eqref{eq:td_target} as the target value.
Even in policy improvement, $y$ is also used when TD error is introduced as a weight.
Thus, $y$ plays a very important role in RL, but little has been said about exception handling at the end of the episode.
Currently, it is standard practice to consider the value after termination to be zero as ``no information''~\cite{sutton2018reinforcement}.
However, is such a zero value really no information?

For example, if the reward is always positive (i.e. $\mathcal{R} \subseteq \mathbb{R}_+$), the value function must be positive.
In this case, it is natural that the value after termination should also be positive, which is ignored in the current exception handling.
Therefore, the value after termination is unintentionally underestimated.
This underestimation implies that the reached endpoint $s^\prime \in \mathcal{S}_\mathrm{term}$ is with a bad result, and encourages exploration to reach other endpoints implicitly.
If the episode is terminated due to task failure, this works effectively to find cases where the task does not fail;
otherwise, if the endpoint implies task success, the agent would not accept this success and continue to find better successful cases.
The same is true when the reward is always negative (i.e. $\mathcal{R} \subseteq \mathbb{R}_-$), and considering the value after termination to be zero unintentionally overestimates the reached endpoints.
This forces the agent to reach the reached endpoints again, even if they imply task failure and are not desired.

As described above, such an implicit bias is easily caused.
Indeed, the behavior according to it can be controlled by properly designing the reward function (for example, simply adding an offset, which remains the optimal policy unchanged~\cite{ng1999policy}) to make the agent promote whether exploitation or exploration.
However, for users who are not familiar with this implicit bias, it can be a fatal issue for utilizing RL in practical applications.
Although an approach to empirically estimate the offset (i.e. the average reward of the task) is conceivable, the implicit bias would remains especially when the estumation accuracy is low.

The above examples are organized in Fig.~\ref{fig:img_under_over}.
It would be natural that the behavior of repeatedly reaching the wrong endpoints due to unintentional overestimation is the most fatal failure, and continuing the exploration even when the agent succeeds in task is allowable in terms of finding better solutions effectively.
Therefore, in this study, a new exception handling is heuristically designed to intentionally underestimate the value after termination without relying on the reward design.
In addition, the degree of underestimation is adjusted according to the degree of stationarity, aiming to suppress unnecessary exploration.

\subsection{Assumption for transition to absorbing state}

\begin{figure}[tb]
    \centering
    \includegraphics[keepaspectratio=true,width=0.96\linewidth]{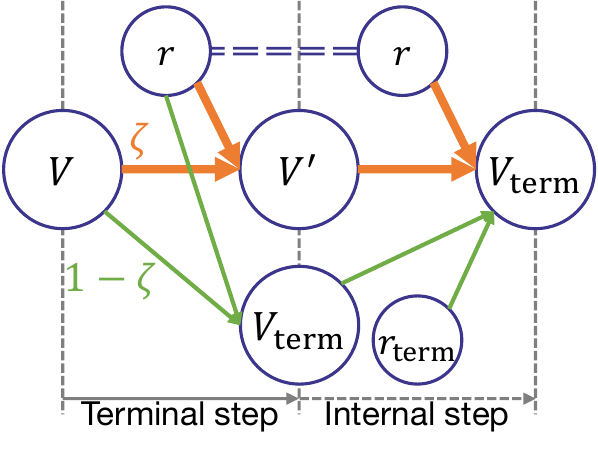}
    \caption{Assumption of the transition to the absorbing state}
    \label{fig:img_trans_absorb}
\end{figure}

In designing the proposed method, the value function after termination is explicitly given as $V_\mathrm{term}$ (or $Q_\mathrm{term}$).
Then, by assuming that the state after termination is regarded to be an absorbing state, the following relationship is given for the reward $r_\mathrm{term}$ obtained in the absorbing state.
\begin{align}
    V_\mathrm{term} = \frac{r_\mathrm{term}}{1 - \gamma}
    \label{eq:value_term}
\end{align}

Next, it is assumed that when the transition is judged to be terminal, a portion of the transitioned state has reached the absorbing state, and the remaining one transits again with the same reward as the previous transition to the fully absorbing state (see Fig.~\ref{fig:img_trans_absorb}).
In other words, the following formulas are assumed according to the recursive structure of the value function.
\begin{align}
    \begin{split}
        V = r + \gamma \{\zeta V^\prime + (1 - \zeta) V_\mathrm{term}\}
        \\
        V^\prime = \zeta r + (1 - \zeta) r_\mathrm{term} + \gamma V_\mathrm{term}
    \end{split}
    \label{eq:trans_assume}
\end{align}
where $V = V(s)$, $V^\prime = V(s^\prime)$ for brevity, and $\zeta \in [0, 1]$ denotes the ratio of $s^\prime$ reaching $V^\prime$ at the first transition.
By substituting eq.~\eqref{eq:value_term} into the second line of eq.~\eqref{eq:trans_assume}, $V_\mathrm{term}$ is derived without using $r_\mathrm{term}$.
\begin{align}
    V_\mathrm{term} = \frac{V^\prime - \zeta r}{1 - \zeta (1 - \gamma)}
    \label{eq:value_term2}
\end{align}

With this, the first line of eq.~\eqref{eq:trans_assume} can be formed with the structure of $r + \gamma V^\prime - U$.
Here, $r + \gamma V^\prime$ is equivalent to $y$ for non-terminated situations, and $U$ means the correction amount at termination.
Specifically, $U$ is given as follows (its derivation process is described in Appendix~\ref{app:derive_U}):
\begin{align}
    U = \frac{\gamma \zeta (1 - \zeta)}{1 - \zeta (1 - \gamma)} \{r - (1 - \gamma) V^\prime\}
    \label{eq:U_normal}
\end{align}

\subsection{Intentional underestimation}

\begin{figure}[tb]
    \centering
    \includegraphics[keepaspectratio=true,width=0.96\linewidth]{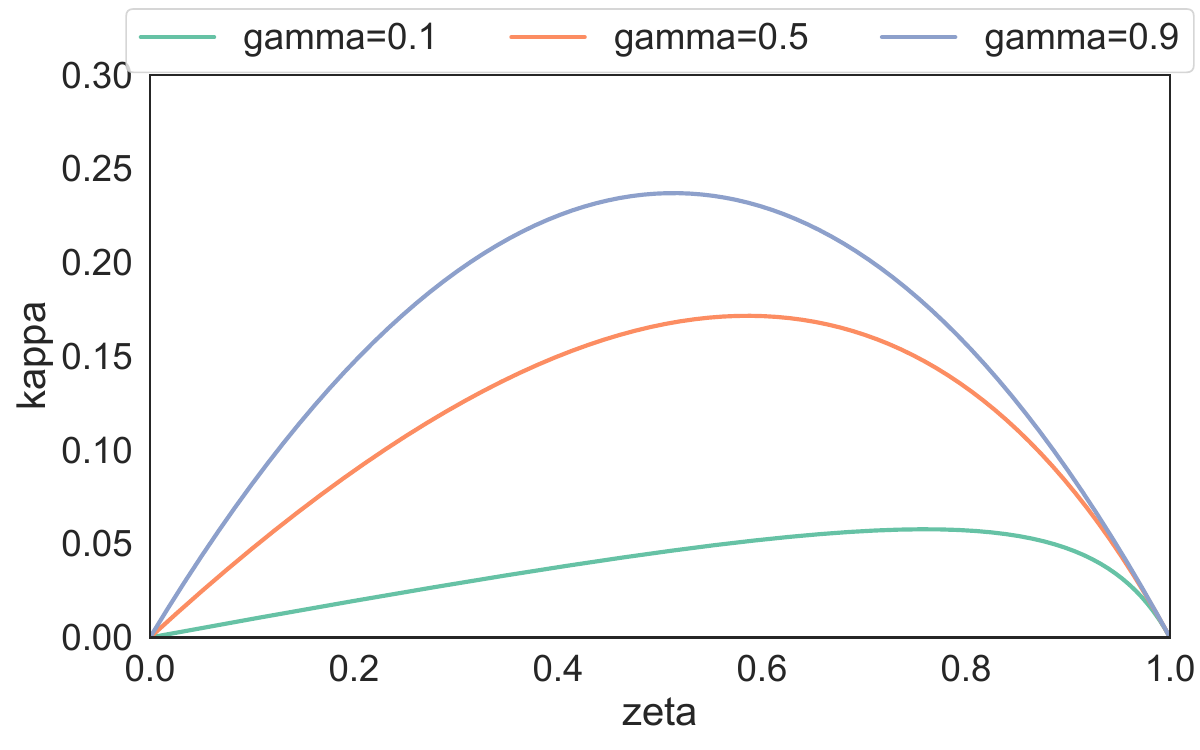}
    \caption{$\zeta$ vs. $\kappa$ with three different $\gamma$}
    \label{fig:img_zeta_kappa}
\end{figure}

If $U$ defined in eq.~\eqref{eq:U_normal} is replaced by a heuristically designed $\tilde{U}$, which satisfies $\tilde{U} \geq U$ and $\tilde{U} \geq 0$, it always yields the intentional underestimation compared to the normal transition at non-terminated situations.
To design $\tilde{U}$, $\frac{\gamma \zeta (1 - \zeta)}{1 - \zeta (1 - \gamma)}$ in eq.~\eqref{eq:U_normal} (defined as $\kappa$) is first designed.
Although $\kappa$ corresponds to the non-negative coefficient, which is determined in accordance with the non-intuitive relation to $\zeta$ (see Fig.~\ref{fig:img_zeta_kappa}), it should be of a certain scale to ensure the effect of underestimation.
Therefore, the maximum value of $\kappa$, $\kappa_{\max}$, is first derived.
The derivation process is described in Appendix~\ref{app:derive_kappa}, and $\kappa_{\max}$ is shown below.
\begin{align}
    \kappa_{\max} = \gamma \left( \frac{1 - \sqrt{\gamma}}{1 - \gamma} \right)^2
    \label{eq:kappa_max}
\end{align}
Based on this, the hyperparameter $\lambda \in [0, 1]$ is multiplied by $\kappa = \lambda \kappa_{\max}$ to intuitively determine the coefficient.
In this paper, $\lambda=0.5$, the center of its domain, is given for simplicity.

Next, the requirements of $\tilde{U}$ is satisfied by modifying the remaining term $r - (1 - \gamma) V^\prime$.
To do so, the following equational transformation is first applied.
\begin{align}
    r - (1 - \gamma) V^\prime &= \gamma (V^\prime - V) - (V^\prime - V)
    \nonumber \\
    &- \gamma (V_r - V) + (V_r - V)
    \nonumber \\
    &= \gamma \Delta V^\prime - \Delta V^\prime - \gamma \Delta V_r + \Delta V_r
    \label{eq:diff_value}
\end{align}
where $V_r = r / (1 - \gamma)$.
Here, two differences, $V^\prime - V = \Delta V^\prime$ and $V_r - V = \Delta V_r$, appear, both of which suggest that the closer they are to zero, the more stationary the state is.
Since $V^\prime$ is a value that may not be learned correctly and $V_r$ is the value assuming that $s$ is already in the absorbing state, both are of uncertain accuracy and unstable.
Therefore, the two differences based on $V$ are introduced.

Anyway, a non-negative upper bound for eq.~\eqref{eq:diff_value} is introduced for $\tilde{U} \geq U$ and $\tilde{U} \geq 0$.
This paper simply makes full use of max/min operators to design it.
\begin{align}
    \eqref{eq:diff_value} &\leq \gamma \max(\Delta V^\prime, 0) - \min(\Delta V^\prime, 0)
    \nonumber \\
    &- \gamma \min(\Delta V_r, 0) + \max(\Delta V_r, 0)
\end{align}
$\tilde{U}$ and $y$ when $s^\prime \in \mathcal{S}_\mathrm{term}$ are finally given as follows:
\begin{align}
    y &= r + \gamma V^\prime - \tilde{U}
    \\
    \tilde{U} &= \lambda \kappa_{\max} \{\gamma \max(\Delta V^\prime, 0) - \min(\Delta V^\prime, 0)
    \nonumber \\
    &\quad\quad\quad\quad - \gamma \min(\Delta V_r, 0) + \max(\Delta V_r, 0)\}
    \label{eq:proposal}
\end{align}

Note that an extreme example of termination with a steady state is exceeding the time limit, and recent implementations of RL benchmark environments have started to distinguish this from other kinds of termination\footnote{This discussion has been raised when developing new APIs on Gymnasium: \url{https://gymnasium.farama.org/}}.
Although the following simulations still handle it as the exception, since the underestimation implemented is suppressed in the steady state, it is expected that the exploration is accelerated only in the early stages of learning, when the steady state is not identified well, and that it can be regarded as no exception thereafter by reducing the underestimation.

\section{Simulations}

\subsection{Simulation conditions}

Numerical simulations are performed to statistically validate the proposed method.
For the simulation environments, Mujoco~\cite{todorov2012mujoco}, Pybullet~\cite{coumans2016pybullet}, and dm\_control~\cite{tunyasuvunakool2020dm_control} are employed with the old APIs in OpenAI Gym.
The proposed method is incorporated into the implementation~\cite{kobayashi2023reward}, which is based on PPO~\cite{schulman2017proximal} with eq.~\eqref{eq:policy_grad1}, and the implementation~\cite{kobayashi2023soft}, which is a modified version of SAC~\cite{haarnoja2018soft} based on eq.~\eqref{eq:policy_grad2}.
The hyperparameters in these implementations remain the same as in the original work.
$\lambda$ in the proposed method is set to $\lambda=0.5$, which is the center of its domain, as described in the above.

The following three comparisons were conducted.
\begin{itemize}
    \item \textit{zero}: The value after termination is assumed to be zero (i.e. the most popular method).
    \item \textit{ignore}: No exception handling at the end of the episode is done.
    \item \textit{underest}: At the end of the episode, eq.~\eqref{eq:proposal} is utilized (i.e. the proposed method).
\end{itemize}
\textit{zero} may fall into a local solution due to the unintended overestimation.
\textit{ignore} is likely to incorrectly estimate the value after termination, which may lead to failure of learning.
In contrast, \textit{underest} should be able to deal with arbitrary reward functions because it can always underestimate the value after termination.

\subsection{Simulation results for positive/negative rewards}

\begin{table*}[tb]
    \caption{Test results for positive/negative rewards: the cases with clearly low performance were underlined.}
    \label{tab:result_pm}
    \centering
    \begin{tabular}{l l | c c c c c c c c}
        \hline\hline
        \multicolumn{2}{l |}{Method} & \multicolumn{8}{c}{Task: mean of return (SD)}
        \\
        && \multicolumn{4}{c}{Pendulum} & \multicolumn{4}{c}{Reacher}
        \\
        && \multicolumn{2}{c}{Mujoco} & \multicolumn{2}{c}{Pybullet} & \multicolumn{2}{c}{Mujoco} & \multicolumn{2}{c}{Pybullet}
        \\
        && Positive & Negative & Positive & Negative & Positive & Negative & Positive & Negative
        \\
        \hline
        & \textit{zero}
        & 5500.9 (4085.7) & \underline{35.6} (0.1) & 8341.4 (2409.0) & \underline{140.5} (3.5)
        & \underline{-12.0} (0.9) & -8.6 (3.0) & \underline{16.6} (1.7) & \underline{1.5} (17.3)
        \\
        \cite{kobayashi2023reward} & \textit{ignore}
        & \underline{1030.8} (2681.2) & \underline{61.2} (24.5) & \underline{1915.8} (3167.4) & \underline{1117.0} (2617.2)
        & -7.2 (1.2) & -7.3 (1.2) & 18.7 (1.1) & 15.3 (4.6)
        \\
        & \textit{underest}
        & 8002.7 (3182.7) & 6019.9 (4171.5) & 5976.2 (2977.5) & 6629.6 (3400.9)
        & -10.3 (1.6) & \underline{-11.4} (1.5) & 18.6 (1.4) & 16.8 (3.3)
        \\
        \hline
        & \textit{zero}
        & 8662.3 (2407.4) & \underline{35.5} (0.0) & 7458.2 (2505.4) & \underline{139.7} (2.7)
        & \underline{-15.2} (2.3) & -6.3 (0.6) & \underline{-5.4} (2.1) & \underline{-11.8} (12.5)
        \\
        \cite{kobayashi2023soft} & \textit{ignore}
        & 9357.1 (3.5) & 6442.9 (4335.1) & 6665.9 (4036.9) & 6256.1 (3939.5)
        & -6.1 (0.7) & -7.3 (1.5) & 17.8 (1.6) & 4.1 (6.4)
        \\
        & \textit{underest}
        & 9358.4 (2.0) & 8132.1 (2877.4) & 9043.3 (1079.8) & 9124.6 (807.7)
        & -7.2 (0.8) & -8.0 (0.9) & 17.8 (2.4) & 3.5 (6.6)
        \\
        \hline\hline
    \end{tabular}
\end{table*}

\begin{table*}[tb]
    \caption{Test results for sparse/complex tasks: the cases with clearly low performance were underlined.}
    \label{tab:result_sc}
    \centering
    \begin{tabular}{l l | c c c c}
        \hline\hline
        \multicolumn{2}{l |}{Method} & \multicolumn{4}{c}{Task: mean of return (SD)}
        \\
        && CartPole (sparse) & FingerSpin (sparse) & Cheetah (complex) & Walker (complex)
        \\
        \hline
        & \textit{zero}
        & \underline{496.1} (372.1) & 158.3 (162.9) & 597.3 (28.6) & \underline{806.2} (275.7)
        \\
        \cite{kobayashi2023reward} & \textit{ignore}
        & 685.8 (219.6) & \underline{135.0} (149.6) & \underline{568.7} (46.1) & 888.4 (141.4)
        \\
        & \textit{underest}
        & 724.9 (85.0) & \underline{133.5} (127.0) & 599.1 (35.7) & 875.5 (213.3)
        \\
        \hline
        & \textit{zero}
        & \underline{59.6} (206.4) & \underline{558.3} (61.3) & \underline{432.7} (21.9) & 957.7 (6.0)
        \\
        \cite{kobayashi2023soft} & \textit{ignore}
        & 134.7 (296.5) & 633.5 (97.7) & 709.2 (41.2) & 968.2 (3.1)
        \\
        & \textit{underest}
        & \underline{66.4} (230.1) & 572.8 (192.3) & 692.6 (42.4) & 955.5 (41.2)
        \\
        \hline\hline
    \end{tabular}
\end{table*}

As toy problems, the following four environments are solved:
Pendulum (i.e. \textit{InvertedDoublePendulum-v4} in Mujoco and \textit{InvertedDoublePendulumBulletEnv-v0} in Pybullet) and; Reacher (i.e. \textit{Reacher-v4} in Mujoco and \textit{ReacherBulletEnv-v0} in Pybullet).
In order to confirm that the behavior of the conventional methods would change significantly depending on the positive or negative reward function, two offsets are added to the original reward function so that it is always either of the positive or negative (see Appendix~\ref{app:offset} for details).

After 2000 episodes of training in each environment and method, the median return for 100 episodes were calculated.
This was done for 12 models with different random seeds, and the statistical performance is summarized in Table~\ref{tab:result_pm}.
Note that the cases with the lower performances than ones with the same conditions except for the exception handling are underlined.

First, the most popular exception handling, \textit{zero}, did not learn the optimal policy at all in Pendulum with the negative rewards, regardless of the algorithm employed.
This is because the exception handling at the task failure unintentionally overestimates the value after termination, repeating the same failure.
In addition, the performance was not sufficiently improved in Reacher with the positive rewards.
This may be due to the fact that even when the agent reached the goal, it continued exploration instead of being satisfied with the result, delaying the convergence of learning.
Reacher on Pybullet failed even with the negative rewards, because it fell into the local solution.
Thus, although \textit{zero} is the most common implementation, it may cause even simple tasks to fail if the rewards are not designed correctly.

Next, focusing on exception-free \textit{ignore}, Pendulum was not solved regardless of reward design in the implementation of the literature~\cite{kobayashi2023reward}, which leverages the value after termination for policy improvement.
This suggests that the implicit exploration, which is facilitated by the exception handling at the end of the episode, was effective to solve Pendulum by policy-gradient-based algorithms.
In fact, Reacher, which may not require exploration so much, succeeded stably.
On the other hand, the implementation of the literature~\cite{kobayashi2023soft}, which uses only the current value for policy improvement, did not fail on any problem, implying the value function was not significantly wrong even with training it using the incorrectly-estimated value after termination.
This is probably because the value function on SAC was trained robustly through the policy entropy maximization and conservatively by selecting the smaller of two approximated value functions.
Thus, depending on the used algorithm, stable learning can be expected without considering the exception handling, although it is difficult to determine which implementations are allowed to do so without being familiar with the algorithm.

Finally, the proposed method \textit{underest} succeeded in stably solving all tasks (including those that could not be solved by \textit{zero} and/or \textit{ignore}), regardless of the algorithm or reward design.
As expected, the proposed method promotes exploration appropriately by always underestimating the value after termination.
In addition, this accelerated exploration seems to be suppressed in the case of Reacher, where the endpoints are stationary, and convergence delay is rarely observed.
The delay is only noticeable when solving Reacher on Mujoco with the negative rewards with the implementation~\cite{kobayashi2023reward}.
It would be possible to solve this issue with a more accurate stationarity evaluation method or automatic adjustment of $\lambda$.
Alternatively, the excessive exploration might be mitigated by ignoring the termination due to time limit as like the recent API.
Thus, although the proposed method leaves room for improvement, it can support successful and stable learning for novices who are unable to design appropriate rewards or select appropriate algorithms.

\subsection{Simulation results for sparse/complex tasks}

As more practical problems, the following four environments with sparse rewards (obtained only when the target tasks are accomplished) or complex dynamics, all of which are implemented in dm\_control~\cite{tunyasuvunakool2020dm_control}, are solved:
CartPole and FingerSpin with sparse rewards and; Cheetah and Walker with complex dynamics.
Even if the reward function is appropriately designed for the target task, performance can vary depending on the exception handling at termination.
Statistical performances were calculated under the same conditions as in the previous simulations and are summarized in Table~\ref{tab:result_sc}.

First, focusing on the results of the algorithm in the literature~\cite{kobayashi2023reward}, it can be found that the proposed method scores high except for FingerSpin.
The low performance on FingerSpin is probably due to insufficient exploration by intentional underestimation, contrary to the aforementioned simulation results.
Nevertheless, the performance of the other two exception handling methods decreased in each of the two tasks, indicating the effectiveness of the proposed method.

On the other hand, with the algorithm based on SAC~\cite{kobayashi2023soft}, the performance of the most popular exception handling \textit{zero} was clearly worse than the others.
The proposed method did not perform well only with CartPole, which may be due to excessive exploration.
Similar to the simulation results described above, the performance of \textit{ignore} was consistently high for this algorithm.

As described above, it was confirmed that even without intentionally skewing the reward in a positive or negative direction, the design of exception handling at termination makes a significant contribution to the learning performance.
While it is clear that the current typical implementation, \textit{zero}, is not always adequate, the proposed method, \textit{underest}, has mostly succeeded in stable learning.
Although there were rare cases of performance degradation due to excessive or insufficient exploration, this could be avoided if the hyperparameter $\lambda$ of the proposed method is appropriately designed.
For example, if $\lambda=0$, the proposed method is consistent with \textit{ignore}, which is desirable for the SAC-based implementation~\cite{kobayashi2023soft}.
Conversely, for the PPO-based implementation~\cite{kobayashi2023reward}, $\lambda > 0.5$ may be better to encourage exploration more.
Alternatively, it would be better to discuss which termination the proposed exception handling should be applied (e.g. ignoring the termination due to time limit).

\section{Experiments}

\subsection{Setup}

\begin{figure}[tb]
    \centering
    \includegraphics[keepaspectratio=true,width=0.96\linewidth]{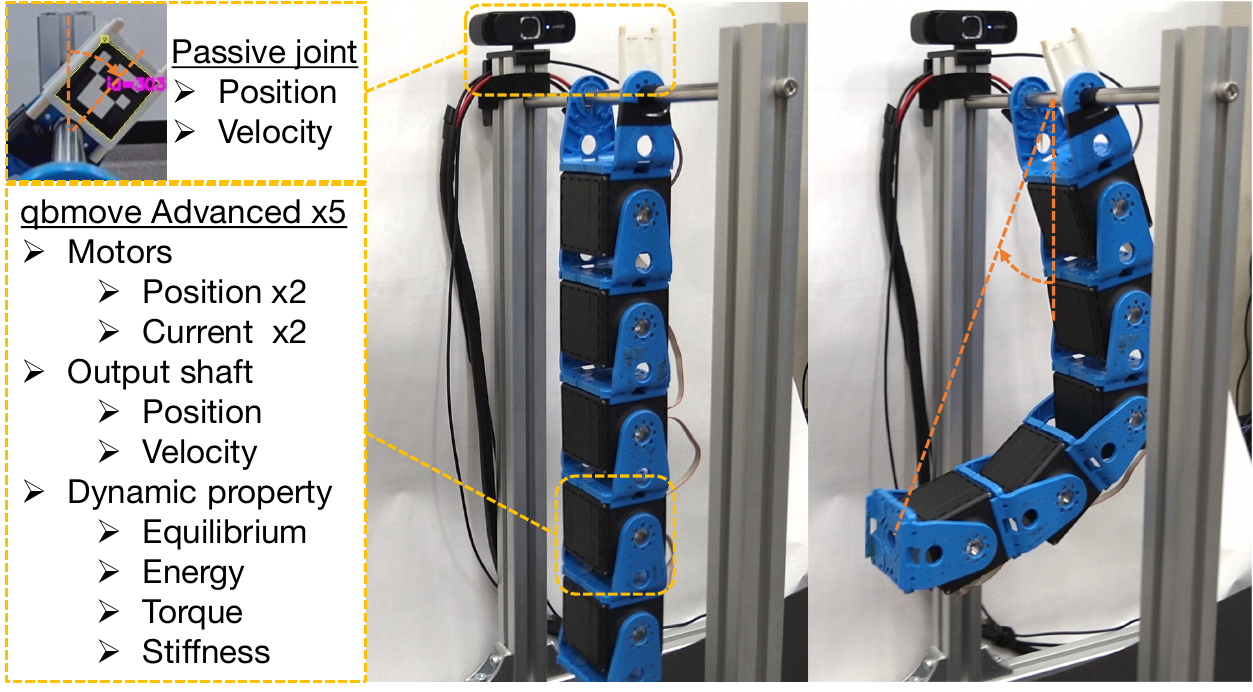}
    \caption{Robot setup}
    \label{fig:exp_robot}
\end{figure}

In the following experiments, only the SAC-based implementation~\cite{kobayashi2023soft} is employed especially for revealing the difference between \textit{ignore} and \textit{underest}, which could not significantly be found in the above simulations.
In addition, the termination due to time limit is not treated as an exception based on the simulation results.

The robot used in the experiments is an underactuated robot with five VSAs~\cite{catalano2011vsa}, which are developed in QbRobotics, connected serially, as shown in Fig.~\ref{fig:exp_robot}.
Each VSA has two built-in motors, and the difference in their outputs makes the stiffness variable.
Therefore, the basic state of each VSA has 10 dimensions: the angles and currents of the two motors; the angle and angular velocity of the output shaft; and the dynamic properties (i.e. equilibrium point, energy, torque, and stiffness).
The action to each VSA is assumed to be two-dimensional: the deviations of the angle and stiffness commands.
Specifically, the action received by each VSA is smoothed and added to the commands, four dimensions of the current commands and the smoothed deviations are added to the state in order to hold MDP.
Since the robot is passively hanging from a bar, the angle and angular velocity around it are detected by a camera and a marker.
In total, this robotic environment contains $14 \times 5 + 2 = 72$ dimensional state space and $2 \times 5 = 10$ dimensional action space.

The target task on this robot is ``synchronized swinging'' like~\cite{hayashibe2022synergetic}.
That is, the angle when the robot is regarded as a pendulum, $q_{\mathrm{pend}}$, is swung as much as possible while all six axes $q_{\mathrm{axes}}$ move in the same direction.
The specific reward function is as follows:
\begin{align}
    r &= - \cos(\max(|q_{\mathrm{pend}}|, |q_{\mathrm{pend}}^\prime|))
    \nonumber \\
    &\times \left( 2 - \cfrac{|\dot{q}_{\mathrm{pend}}^\prime|}{2\pi} \right)
            \left( 2 - \cfrac{|\sum_i \dot{q}_{\mathrm{axes},i}^\prime|}{\sum_ie |\dot{q}_{\mathrm{axes},i}^\prime|} \right)
\end{align}
where the first term is set for amplitude, the second for frequency, and the third for synchronization.
This reward is negative due to $|q_{\mathrm{pend}}| < \pi/2$ and $|\dot{q}_{\mathrm{pend}}| < 4\pi$.
That is, for \textit{zero}, the terminal state at the task failure shown below is overestimated.

Here, the task terminates when the maximum current consumption among all motors exceeds 1~A before and after the transition as failure.
Since the current consumption is not involved in the above reward function, \textit{ignore} is prone to repeat the same failure without understanding why the task was terminated.
In contrast, \textit{underest} can explore/discover behaviors that avoid it by intentionally underestimating the value at failure.
Note that the time limit is given as 300 steps (approximately 15 seconds), although no exception handling is applied for this as mentioned above.

\subsection{Results}

\begin{figure}[tb]
    \centering
    \includegraphics[keepaspectratio=true,width=0.96\linewidth]{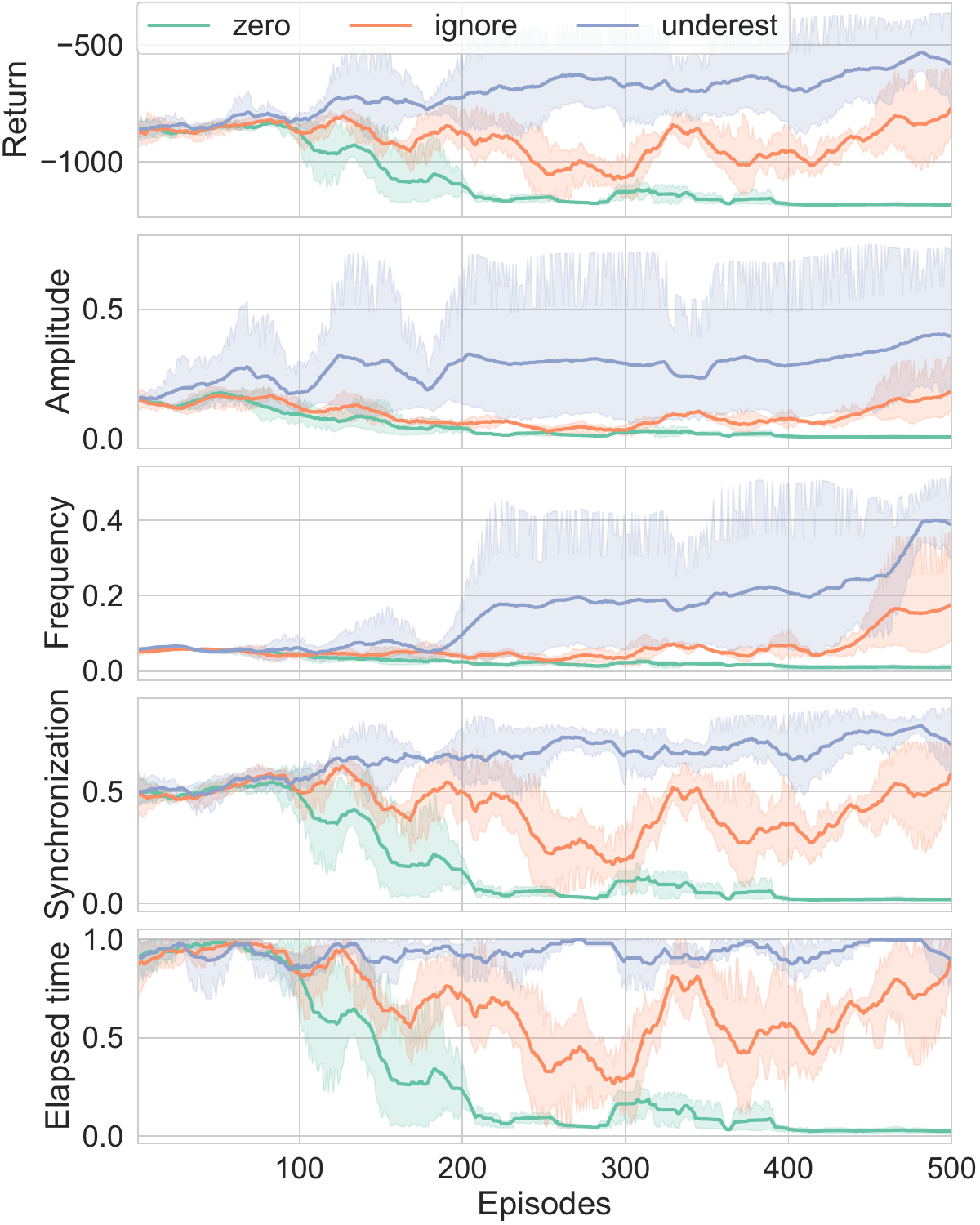}
    \caption{Learning curves of the synchronized swinging}
    \label{fig:exp_learn}
\end{figure}

\begin{figure}[tb]
    \centering
    \includegraphics[keepaspectratio=true,width=0.96\linewidth]{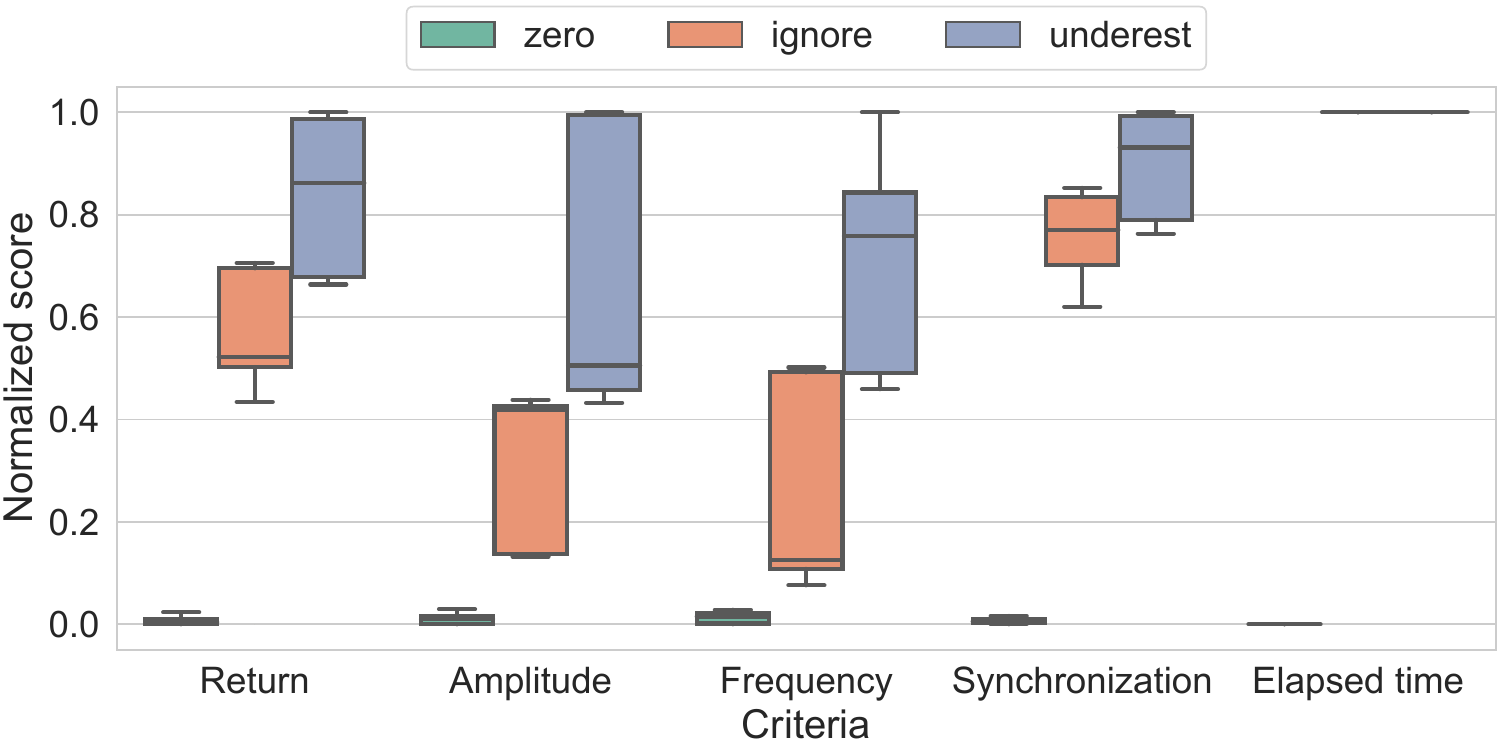}
    \caption{Test performance of the synchronized swinging}
    \label{fig:exp_test}
\end{figure}

Three trials were conducted for each method with different seeds.
The learning curves for the return as well as the three terms involved and the rate of time elapsed until the end of the episode (the larger the better) are shown in Fig.~\ref{fig:exp_learn}.
Note that when teh task fails, it was evaluated as if the lowest value of each criterion was obtained in the remaining time steps.
The results of five tests by the trained policies are summarized in Fig.~\ref{fig:exp_test}.
A part of them can be seen in the attached video.

In all methods, after a while from the start of learning, the current consumption of the motors became too large for swinging, and the task failed due to overcurrent.
The subsequent response differed among the methods.
As expected, \textit{zero} overestimated the value at the failure, resulting in convergence to the failure state as fast as possible.
In the case of \textit{ignore}, the agent tried to maximize the return without knowing the cause of the failure, which led to repeated failures without suppressing the current consumption.
As a result, the number of samples per episode was reduced, deteriorating the learning efficiency.

In contrast, \textit{underest} underestimated the value at the failure to avoid making it.
Thus, it successfully maximized all the criteria while the current consumption was suppressed.
Indeed, as shown in the attached video, only \textit{underest} could achieve the synchronized swinging.

\section{Conclusion}

This paper pointed out the issue with the current mainstream for treating the value after termination as zero with regard to exception handling at termination in RL with TD learning.
A new implementation was then proposed to always implicitly encourage exploration by intentionally underestimating the value after termination, regardless of what kind of reward function is designed.
Numerical simulations and real robot experiments indicated that the proposed method succeeded in learning even with mis-designed rewards where the conventional methods failed.

However, the design of the proposed method includes heuristics, and there is room for improvements.
As one of the considerable future improvements, more detailed exception conditions would be explicitly required so that a new method (based on the proposed method) handles the respective exceptions more appropriately.

\appendix

\subsection{Derivation of eq.~\eqref{eq:U_normal}}
\label{app:derive_U}

By substituting eq.~\eqref{eq:value_term2} into the first line of eq.~\eqref{eq:trans_assume}, the following formula is hold.
\begin{align*}
    V &= r + \gamma \left\{ \zeta V^\prime + (1 - \zeta) \frac{V^\prime - \zeta r}{1 - \zeta (1 - \gamma)} \right\}
    \\
    &= r + \gamma V^\prime - \gamma (1 - \zeta) V^\prime + \frac{\gamma (1 - \zeta)}{1 - \zeta (1 - \gamma)} (V^\prime - \zeta r)
    \\
    &= r + \gamma V^\prime - \gamma (1 - \zeta) \left\{ V^\prime - \frac{1}{1 - \zeta (1 - \gamma)} (V^\prime - \zeta r) \right\}
    \\
    &= r + \gamma V^\prime - \frac{\gamma (1 - \zeta)}{1 - \zeta (1 - \gamma)} \{ -\zeta (1 - \gamma) V^\prime + \zeta r \}
    \\
    &= r + \gamma V^\prime - \frac{\gamma \zeta (1 - \zeta)}{1 - \zeta (1 - \gamma)} \{r - (1 - \gamma V^\prime)\}
\end{align*}
The last term is $U$ and coincides with eq.~\eqref{eq:U_normal}.

\subsection{Derivation of eq.~\eqref{eq:kappa_max}}
\label{app:derive_kappa}

Since $\kappa$ is non-negative and unimodal to $\zeta$, solving $\partial \kappa / \partial \zeta = 0$ leads to $\zeta_{\max}$, which yields $\kappa_{\max}$.
Specifically, $\partial \kappa / \partial \zeta = 0$ is derived as follows:
\begin{align*}
    \frac{\partial \kappa}{\partial \zeta} &= \frac{\gamma (1 - 2 \zeta)\{1 - \zeta (1 - \gamma)\} + \gamma (1 - \gamma) \zeta (1 - \zeta)}{\{1 - \zeta (1 - \gamma)\}^2}
    \\
    &\propto (1 - 3\zeta + \zeta \gamma +2 \zeta^2 (1 - \gamma)) + \gamma (1 - \gamma) (\zeta - \zeta^2)
    \\
    &= \gamma (1 - 2\zeta + (1 - \gamma) \zeta^2)
    \\
    &\propto (1 - \gamma) \zeta^2 - 2 \zeta + 1 = 0
\end{align*}
where the proportional symbol means that unnecessary nonzero terms are removed.
Since it is now a quadratic equation about $\zeta$, its solution can be found in the following one under the domain $\zeta \in [0, 1]$.
\begin{align*}
    \zeta_{\max} = \frac{1 \pm \sqrt{1 - (1 - \gamma)}}{1 - \gamma}
    = \frac{1 - \sqrt{\gamma}}{1 - \gamma}
\end{align*}

By substituting $\zeta_{\max}$ (or $\zeta_{\max} (1 - \gamma) = 1 - \sqrt{\gamma}$) into $\kappa$, $\kappa_{\max}$ is derived as follows:
\begin{align*}
    \kappa_{\max} &= \frac{\gamma}{1 - \zeta_{\max}(1 - \gamma)} \zeta_{\max} (1 - \zeta_{\max})
    \\
    &= \frac{\gamma}{\sqrt{\gamma}} \frac{1 - \sqrt{\gamma}}{1 - \gamma} \left( \frac{(1 + \sqrt{\gamma})(1 - \sqrt{\gamma})}{1 - \gamma} - \frac{1 - \sqrt{\gamma}}{1 - \gamma} \right)
    \\
    &= \frac{\gamma}{\sqrt{\gamma}} \frac{1 - \sqrt{\gamma}}{1 - \gamma} \frac{\sqrt{\gamma}(1 - \sqrt{\gamma})}{1 - \gamma}
    \\
    &= \gamma \left( \frac{1 - \sqrt{\gamma}}{1 - \gamma} \right)^2
\end{align*}

\subsection{Offsets for positive/negative rewards}
\label{app:offset}

Pendulum is originally with the positive reward function in both Mujoco and Pybullet.
Therefore, the offset for the positive reward function is zero; and the originally maximum reward, 10, is subtracted as the offset for the negative reward function.
Reacher in Mujoco is originally with the negative reward function, so 10 is added as the offset for the positive reward function; and the case for the negative reward function has no change.
On the other hand, Reacher in Pybullet is originally designed to get the positive and negative rewards, hence, 10 is added/subtracted for the positive/negative reward functions, respectively.
Note again that these offsets should not change the optimal policy in theory of reward shaping~\cite{ng1999policy}.

\section*{ACKNOWLEDGMENT}

This work was supported by JSPS KAKENHI, Grant-in-Aid for Scientific Research (B), Grant Number JP21H01353.

\bibliographystyle{IEEEtran}
{
\bibliography{biblio}
}


\end{document}